\documentclass[10pt,journal,compsoc]{IEEEtran}
\newif\ifpeerreview
\peerreviewfalse

\usepackage[nocompress]{cite}
\usepackage{url}
\usepackage{amsmath,amssymb,graphicx}
\usepackage{algorithm}
\usepackage{algpseudocode}
\usepackage{booktabs}
\usepackage{multirow}
\usepackage{capt-of}
\usepackage{changepage}
\usepackage[caption=false,font=footnotesize]{subfig}
\usepackage[table]{xcolor}
\usepackage[switch]{lineno}

\definecolor{tablecolor}{RGB}{226,236,218}

\makeatletter
\def\footnoterule{\relax\if@IEEEenableoneshotfootnoterule
  \kern-5pt
  \hbox to \columnwidth{\vrule width 0.5\columnwidth height 0.4pt\hfill}
  \kern4.6pt
  \global\@IEEEenableoneshotfootnoterulefalse
\else
  \relax
\fi}
\makeatother

\newcommand{\paperID}{43}
\title{Towards Redundancy Reduction in Diffusion Models for Efficient Video Super-Resolution}
\author{Jinpei Guo$^{1,2}$, Yifei Ji$^{2\ddagger}$, Shengwei Wang$^{3\ddagger}$, Zheng Chen$^{2}$, Yufei Wang$^{4}$,\\
Sizhuo Ma$^{4}$, Yong Guo$^{5}$, Baiang Li, Jusheng Zhang$^{3}$,\\
Yulun Zhang$^{2*\dagger}$, Jian Wang$^{4*}$\\
$^{1}$Carnegie Mellon University, $^{2}$Shanghai Jiao Tong University, $^{3}$Sun Yat-sen University,\\
$^{4}$Snap Inc., $^{5}$South China University of Technology%
\thanks{\normalfont$^{\ddagger}$Equal contribution.\protect\\
$^{*}$Equal advising.\protect\\
$^{\dagger}$Corresponding author: Yulun Zhang, yulun100@gmail.com}}

\newcommand{\eg}{\emph{e.g.}}
\newcommand{\ie}{\emph{i.e.}}

\providecommand{\citep}[1]{\cite{#1}}
\providecommand{\citet}[1]{\cite{#1}}

\begin{document}
\setlength{\abovedisplayskip}{2pt}
\setlength{\belowdisplayskip}{2pt}

\IEEEtitleabstractindextext{%
\begin{abstract}
Diffusion models have recently shown promising results for video super-resolution (VSR). However, directly adapting these generative models to VSR creates a task mismatch and results in redundancy, since low-quality videos already preserve substantial content information. Such redundancy leads to increased computational overhead and learning burden, as the model performs superfluous operations and must learn to filter out irrelevant information. To address this problem, we propose OASIS, an efficient $\textbf{o}$ne-step diffusion model with $\textbf{a}$ttention $\textbf{s}$pecialization for real-world v$\textbf{i}$deo $\textbf{s}$uper-resolution. OASIS incorporates an attention specialization routing that assigns attention heads to different patterns according to their intrinsic behaviors while replacing the heavy VAE encoder with a pixel-unshuffle operation followed by a linear projection and omitting the prompt extractor. This adaptation mitigates redundancy while effectively preserving pretrained knowledge, allowing diffusion models to better adapt to VSR and achieve stronger performance. Moreover, to ease the learning burden caused by these architectural simplifications, we design a simple yet effective progressive training strategy, where training starts with temporally consistent degradations and then shifts to inconsistent settings. This strategy significantly facilitates learning under complex degradations. Extensive experiments demonstrate that OASIS achieves state-of-the-art performance on both synthetic and real-world datasets. OASIS also provides superior inference speed, offering a $\textbf{6.2$\times$}$ speedup over one-step diffusion baselines such as SeedVR2. The code and models will be publicly available.
\end{abstract}

}

\maketitle

\IEEEraisesectionheading{
  \section{Introduction}\label{sec:introduction}
}
\begin{figure*}[t]
    \centering
    \includegraphics[width=\textwidth]{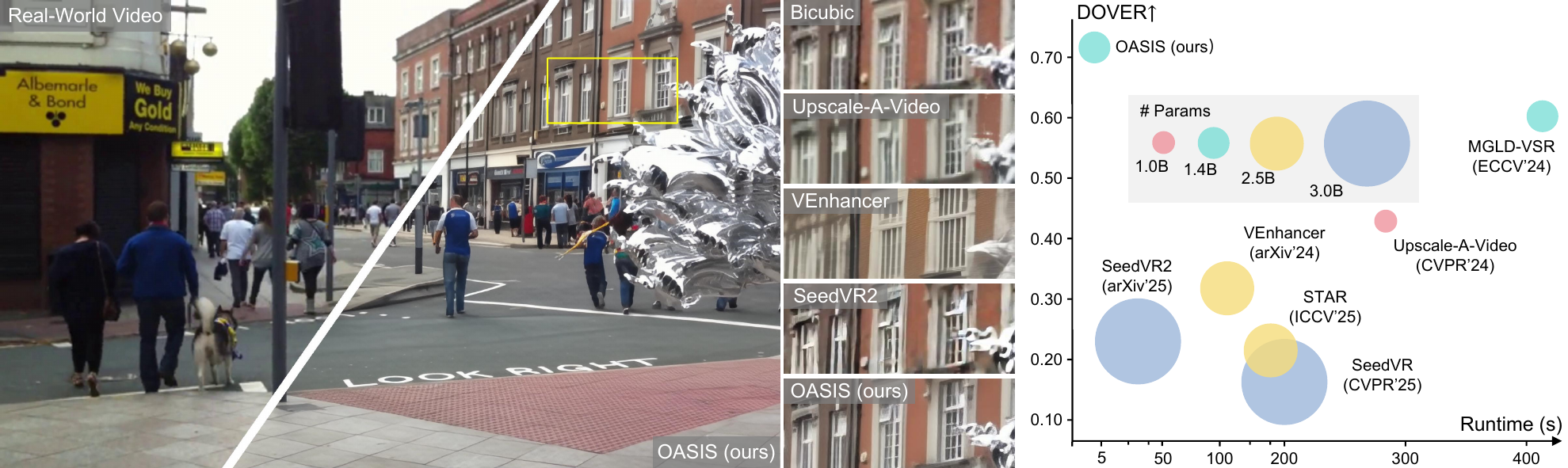}
    \caption{Inference speed and performance comparisons. The running time is evaluated on an A100 GPU using a 33-frame 720$\times$1280 video, while DOVER is reported on the MVSR4x dataset. Our OASIS demonstrates superior reconstruction quality over existing diffusion-based methods, producing clearer and more faithful details. At the same time, OASIS also provides higher inference efficiency. Compared with SeedVR2~\citep{wang2025seedvr2}, it runs approximately 6.2$\times$ faster.}
    \vspace{-4mm}
    \label{fig:teaser}
\end{figure*}

Video super-resolution (VSR) is a widely studied task that aims to reconstruct high-resolution videos from low-resolution inputs~\citep{jo2018deep}. With the explosive growth of social media, enhancing real-world videos has become increasingly important. In contrast to synthetic degradations, such as bicubic downsampling, real-world videos often undergo far more diverse and unpredictable degradations, including varying levels of blur, noise, and compression artifacts~\citep{chan2021basicvsr, wang2021real}. These complex conditions substantially increase the difficulty of restoring accurate and temporally consistent high-resolution content, making real-world VSR a challenging problem.

Recent diffusion models~\citep{ho2020denoising, song2020denoising}, especially diffusion transformer (DiT) architectures~\citep{peebles2023scalable, ma2024sit}, have demonstrated remarkable image and video generation capabilities~\citep{rombach2022high, yang2024cogvideox, wan2025wan}. These models learn rich visual priors and strong spatio-temporal representations, making them appealing foundations for video restoration~\citep{zhou2024upscale, li2025diffvsr, du2025patchvsr}. One-step diffusion further reduces the number of sampling steps~\citep{liu2025ultravsr, wang2025seedvr2}, suggesting a promising path toward practical diffusion-based VSR.

However, adapting a video generative diffusion model to VSR is not simply a matter of reusing a stronger backbone. Video generation starts from noise and must synthesize both structure and detail, whereas VSR is a conditioned restoration task: the low-quality input already provides coarse geometry, motion trajectories, and much of the scene content. This difference creates a task mismatch. If the full generative machinery is retained without modification, the model may spend computation on information that is already available in the input, and the finetuning process must learn to suppress irrelevant long-range interactions rather than focusing on restoration.

Existing diffusion-based VSR methods often address adaptation by adding extra conditioning or temporal modules, such as temporal layers~\citep{zhou2024upscale, li2025diffvsr}, or optical-flow networks~\citep{yang2024motion}. These designs can improve restoration quality, but they also increase complexity. Other methods attempt larger architectural redesigns~\citep{wang2025seedvr2, wang2025seedvr}, which may disturb pretrained knowledge and require expensive retraining (\eg, 256 H100-80G GPUs). This motivates a different question: instead of adding more machinery to a pretrained video diffusion model, can we identify which parts are redundant for VSR and adapt them in a task-specific manner?

We begin from the attention mechanism in DiTs. Although video diffusion models commonly apply global attention uniformly to all heads~\citep{kondratyuk2023videopoet, yang2024cogvideox, kong2024hunyuanvideo, wan2025wan}, our analysis shows that many heads already behave as localized processors across different videos. Besides genuinely global heads, two stable local patterns emerge: intra-frame attention, which mainly captures dependencies within the current frame, and window attention, which focuses on local spatio-temporal neighborhoods. This head-level specialization suggests that uniform global attention is over-complete for VSR, thereby introducing redundancy. In conditioned restoration, a larger receptive field is not always more useful; unnecessary long-range interactions can dilute local texture and motion cues that are already present in the input.

Motivated by this observation, we propose OASIS, an efficient \textbf{o}ne-step diffusion model with \textbf{a}ttention \textbf{s}pecialization for real-world v\textbf{i}deo \textbf{s}uper-resolution. OASIS adapts a pretrained video diffusion transformer by routing attention heads to global, intra-frame, or window patterns according to their intrinsic attention distributions. Instead of treating all heads as identical global processors, this routing is determined by computing the KL-divergence between the original global attention distributions and localized alternatives. Attention heads that align more closely with localized patterns are reassigned to the corresponding modes, while the rest retain global attention. This routine mitigates redundancy while effectively preserving pretrained knowledge, allowing diffusion models to better adapt to VSR.

Beyond attention, OASIS also addresses redundancy in conditioning components. Since the LQ video itself preserves substantial structural information, routing it through a heavy VAE encoder and a prompt extractor introduces unnecessary computational cost. We therefore replace the redundant VAE encoder with a lightweight pixel-unshuffle projection and omit the prompt extractor. However, these VSR-oriented architectural reductions---altering attention receptive fields and removing encoders---further complicate the learning process if the model is directly trained under complex, unpredictable degradations. To address this, we introduce a simple yet effective progressive training strategy. Specifically, in the first stage, the model is trained with temporally consistent degradations to learn fundamental restoration capabilities. In the second stage, the training shifts to temporally inconsistent degradations, where each frame undergoes frame-wise varying distortions. This curriculum strategy reduces the learning burden, and encourages the model to handle frame-wise variations, thereby improving robustness. As shown in Fig.~\ref{fig:teaser}, OASIS achieves strong reconstruction quality and substantially faster inference.

Our main contributions are summarized as follows:

\begin{itemize}
\item We propose a novel and efficient one-step diffusion model, OASIS, for real-world VSR. By incorporating an attention specialization routing, along with replacing the VAE encoder with a lightweight pixel-unshuffle operation followed by a linear projection and omitting the prompt extractor, OASIS mitigates the redundancy of pretrained diffusion transformers when adapted to VSR, thereby reducing computational cost and making the model better leverage diverse attention patterns for high-quality restoration.
\item We design a simple yet effective progressive training strategy, where the model is first trained with temporally consistent degradations and then with temporally inconsistent settings to better reflect real-world scenarios. This strategy reduces the learning burden caused by architectural simplifications, enabling the model to handle complex degradations more effectively.
\item OASIS achieves state-of-the-art results on multiple benchmarks, excelling in both quantitative metrics and perceptual quality. Moreover, it provides remarkable inference efficiency compared with existing diffusion-based VSR methods.
\end{itemize}

\begin{figure*}[t]
    \begin{center}
        \includegraphics[width=\textwidth]{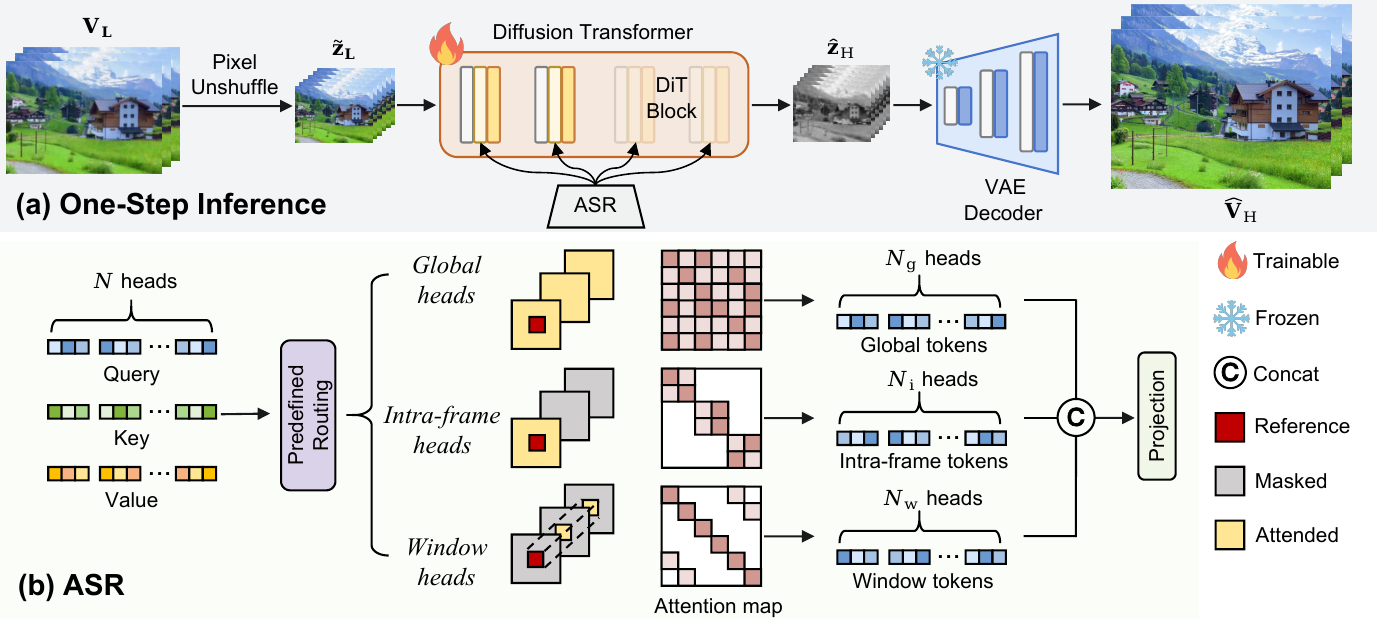}
    \end{center}
    \vspace{-3mm}
    \caption{Overview of OASIS. Given an input LQ video, a pixel-unshuffle operation maps it into the latent space, which is then processed by a diffusion transformer with attention specialization routing (ASR). ASR reduces redundancy by dividing attention heads into global, intra-frame, and window groups to capture complementary contexts. Their outputs are concatenated into an aggregated feature, and a VAE decoder reconstructs the HQ video from the restored latent.} 
    \vspace{-3mm}
    \label{fig:pipeline}
\end{figure*}

\section{Related Work}
\subsection{Video Super-Resolution}
In recent years, learning-based approaches have driven significant progress in video super-resolution (VSR)~\citep{isobe2020video, chan2021basicvsr, chan2022basicvsr++, li2023simple, chen2024learning}. These methods exploit diverse architectures, ranging from deformable convolutions~\citep{wang2019edvr, tian2020tdan} to transformer-based designs~\citep{li2020mucan, liang2022recurrent, shi2022rethinking}. Inspired by the success of GANs in image restoration, several GAN-based methods~\citep{lucas2019generative, xu2025videogigagan} have also been introduced to recover fine-grained details.

Despite these advances, existing methods often struggle under complex real-world degradations. To enhance robustness, some works~\citep{yang2021real, wang2023benchmark} leverage real-world LQ–HQ paired data to enhance robustness. Others focus on architectural redesigns~\citep{pan2021deep, wu2022animesr, zhang2024realviformer} to improve adaptability to challenging degradations. In parallel, various degradation pipelines have been proposed to better simulate real-world conditions~\citep{wang2021real, chan2022investigating}. Nevertheless, these methods still exhibit limited performance when faced with diverse and unpredictable real-world degradations.

\subsection{Diffusion Models}
Diffusion models are powerful generative frameworks that synthesize structured data from random noise via iterative denoising~\citep{ho2020denoising, song2020denoising}. Recently, they have achieved strong performance in both image~\citep{rombach2022high, podell2023sdxl} and video~\citep{ho2022video, zheng2024open, yang2024cogvideox, wan2025wan} generation. However, multi-step diffusion models are often hindered by slow inference, motivating the development of one-step approaches for acceleration~\citep{liu2022flow, song2023consistency, yin2024one, lin2025diffusion}.

Owing to the strong generative prior, diffusion models have also shown competitive performance in image and video restoration~\citep{zhou2024upscale, guo2025compression, guo2025oscar, li2025diffvsr, wang2025seedvr}. For instance, Upscale-A-Video~\citep{zhou2024upscale} extends image diffusion models with temporal layers for video sequences, while MGLD-VSR~\citep{yang2024motion} leverages optical flow to refine latent sampling for better temporal coherence. STAR~\citep{xie2025star} incorporates a local enhancement module to restore fine details, and SeedVR~\citep{wang2025seedvr} employs a sliding-window strategy to handle long sequences. More recently, several works have explored one-step acceleration for faster inference (\eg, SeedVR2~\citep{wang2025seedvr2};~\citep{liu2025ultravsr, sun2025one}). However, most existing approaches overlook the inherent redundancy in pretrained diffusion models, which limits their effectiveness when directly adapted to VSR.

\subsection{Redundancy Reduction in Diffusion Models}
Recent studies have highlighted that diffusion models, despite their strong generative power, often suffer from substantial redundancy~\citep{sun2024unveiling, zhang2023dimensionality, zhang2025blockdance, zhao2024dynamic}, which becomes especially pronounced in restoration tasks~\citep{chen2025adversarial} since low-quality videos already contain much of the underlying content. To address this, a growing body of work has focused on reducing redundancy to improve efficiency~\citep{castells2024edgefusion, zhu2023sdm, zhang2024effortless, pu2024efficient, sun2024asymrnr, tian2025dic}.

In particular, attention redundancy in DiTs has attracted considerable interest. DiTFastAttn~\citep{yuan2024ditfastattn} reduces redundant computation through attention sharing and localized attention patterns. SVG~\citep{xi2025sparse} leverages the inherent sparsity of 3D spatiotemporal attention by profiling head types and applying sparse patterns with kernel optimizations, while STA~\citep{zhang2025fast} eliminates redundancy from global attention with a hardware-aware sliding window design. However, VSR presents different attention characteristics from video generation, and there is still no systematic study on how to reduce attention redundancy in diffusion-based VSR models while achieving consistent performance gains.

\section{Method}
\subsection{Preliminaries: One-Step Diffusion Model}
Latent Diffusion Models ~\citep{rombach2022high} are formulated in a low-dimensional latent space, leading to improved efficiency in training and sampling. In the forward process, a clean latent $\mathbf{z}_0$ is gradually perturbed by Gaussian noise. At timestep $t$, the corrupted latent can be written as:
\begin{equation}
\mathbf{z}_t =\alpha_t\mathbf{z}_0 + \sigma_t\boldsymbol{\epsilon}, \quad \boldsymbol{\epsilon}\sim \mathcal{N}(\mathbf{0}, \mathbf{I}),
\end{equation}
where $\alpha_t$ and $\sigma_t$ are the noise schedule conditioned on the timestep. 
The reverse process then reconstructs the clean latent through a learned prediction model~\citep{ho2020denoising}, where transformer-based architectures~\citep{peebles2023scalable, ma2024sit} have demonstrated strong performance. 
In the one-step setting, the network $\boldsymbol{\epsilon}_\theta$ directly estimates the clean latent $\hat{\mathbf{z}}_0$ from the noised latent:
\begin{equation}
\hat{\mathbf{z}}_0 = \left(\mathbf{z}_t - \sigma_t\boldsymbol{\epsilon}_\theta(\mathbf{z}_t,t)\right)/\alpha_t.
\end{equation}

\subsection{Overview of OASIS}
An overview of OASIS is shown in Fig.~\ref{fig:pipeline}, which is built upon Wan2.1~\citep{wan2025wan}, a powerful pretrained text-to-video diffusion model. Following prior work~\citep{chen2025adversarial}, we omit the VAE encoder to avoid redundant encoding and instead apply a pixel-unshuffle operation~\citep{shi2016real} followed by a linear projection to directly map the input LQ video $\mathbf{V}_{\text{L}}$ into the latent space:
\begin{equation}
    \tilde{\mathbf{z}}_{\text{L}}=\operatorname{UnShuffle}(\mathbf{V}_{\text{L}}), \quad \mathbf{z}_{\text{L}}=\mathbf{W}_{\text{proj}}\tilde{\mathbf{z}}_{\text{L}}+\mathbf{b}_{\text{proj}},
\end{equation}
where $\mathbf{W}_{\text{proj}}$ and $\mathbf{b}_{\text{proj}}$ are the parameters of the linear projection layer. Unlike standard diffusion models that start from Gaussian noise, OASIS views LQ latents $\mathbf{z}_{\text{L}}$ as intermediate diffusion states and their high-quality (HQ) counterparts $\mathbf{z}_{\text{H}}$ as the clean target state. Since Wan2.1 adopts the flow matching formulation, the reconstruction can thus be formulated as:
\begin{equation}
    \hat{\mathbf{z}}_{\text{H}} = \mathbf{z}_{\text{L}} - \sigma_{T_{\text{L}}}\mathcal{DN}_\theta(\mathbf{z}_{\text{L}},{T_{\text{L}}}),
\end{equation}
where $\hat{\mathbf{z}}_{\text{H}}$ is the estimated high-quality video latents, $\mathcal{DN}_\theta$ is the DiT integrated with our proposed attention specialization routing (ASR), and $T_{\text{L}}$ is the predefined timestep. Finally, the reconstructed HQ video $\hat{\mathbf{V}}_{\text{H}}$ is decoded from $\hat{\mathbf{z}}_{\text{H}}$ using the 3D VAE decoder $\mathcal{D}_\phi$: $\hat{\mathbf{V}}_{\text{H}}=\mathcal{D}_\phi\left(\hat{\mathbf{z}}_{\text{H}}\right)$.
\begin{figure}[h]

    \centering
    \includegraphics[width=\linewidth]{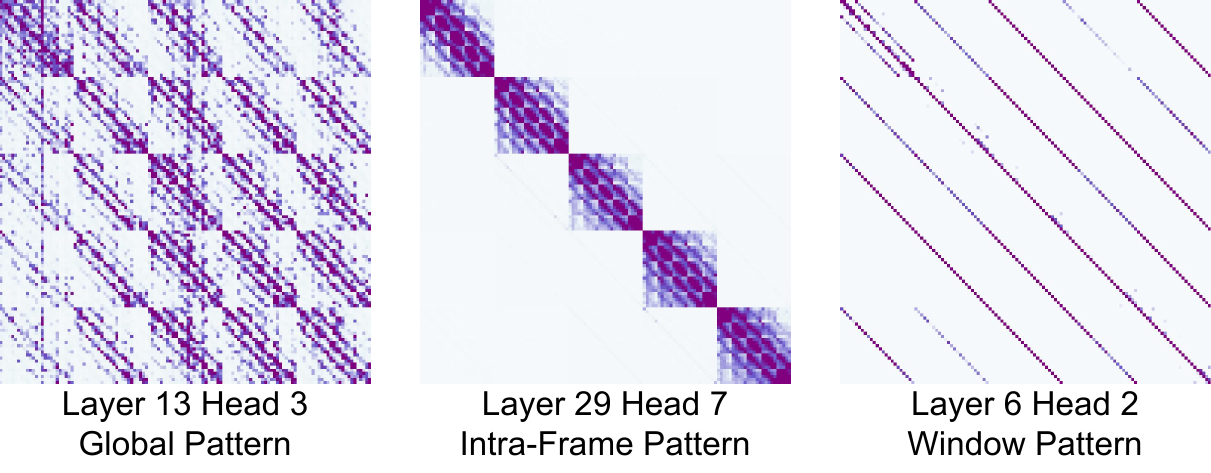}
    \vspace{-6mm}
    \caption{Head-level specialization in DiTs. Each visualization is the mean attention map over 50 videos from the training set, revealing global, intra-frame, and window patterns. }
    \label{fig:attn_map}
    \vspace{-5mm}
\end{figure}
\subsection{Redundancy Reduction}
\subsubsection{Attention Redundancy in Diffusion Transformers}
When video generative diffusion models are adapted to VSR, they often introduce redundancy, as low-quality videos already retain content information. To address this, we investigate redundancy patterns in video generative diffusion models and find that attention is one common source.

As shown in Fig.~\ref{fig:attn_map}, we obtain the visualization by running inference on 50 videos and averaging the attention maps for each head. We find that the attention pattern of each head remains highly consistent across different inputs. Since the maps produced by the same head look nearly identical for all videos, the averaged map naturally preserves the same pattern. The results clearly show that while certain heads present global behavior, a subset of heads consistently display localized behavior, such as intra-frame attention or window attention. These visualizations highlight that, although formulated as global attention, not all heads truly exploit global context, revealing clear redundancy in uniform global attention assignments. We provide more visualizations in the Appendix.

\begin{figure*}[t]
    \centering
    \includegraphics[width=0.9\textwidth]{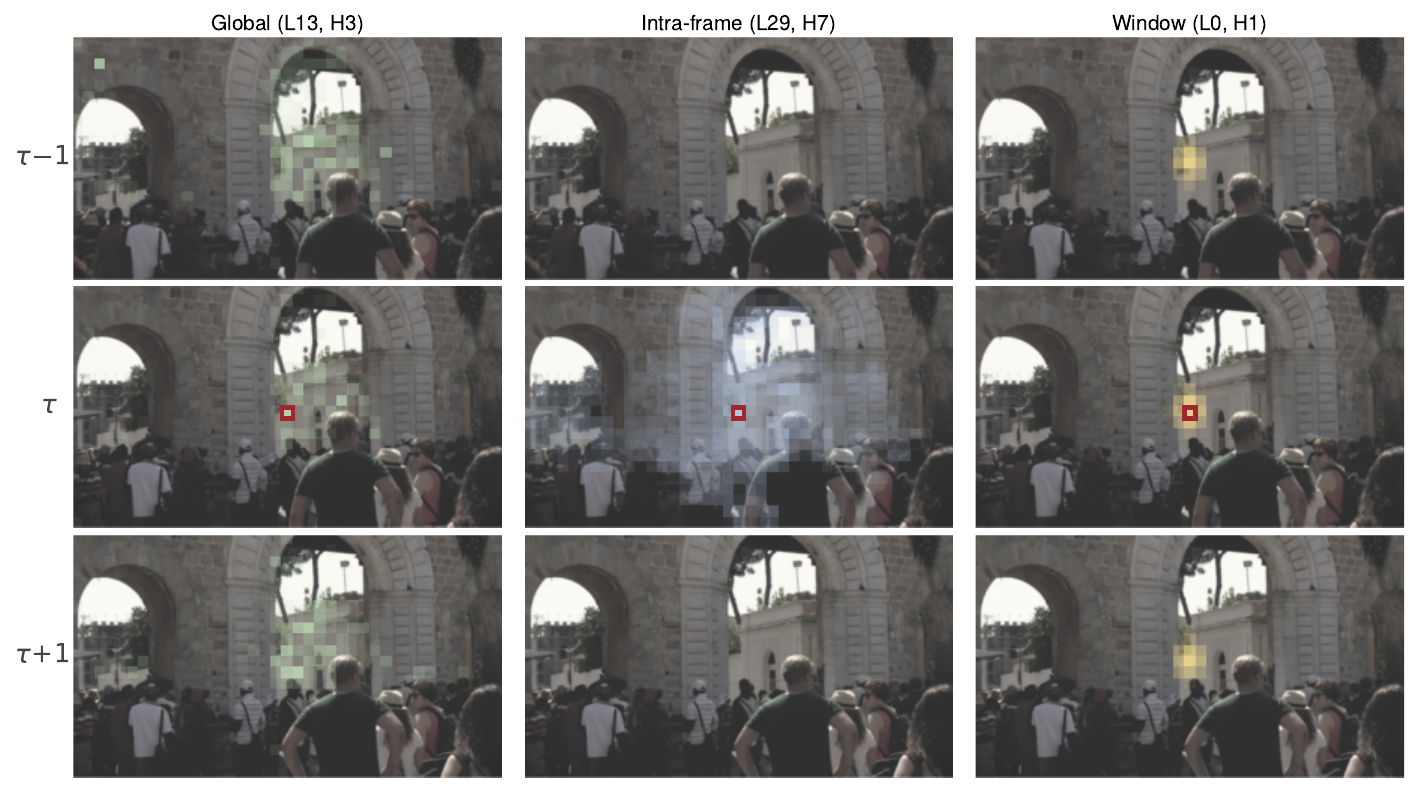}
    \vspace{-3mm}
    \caption{Application of routed attention patterns on real video frames. A fixed query token is highlighted on the middle slice. Colored overlays show where each routed head places attention. The global head attends broadly across both space and time, the intra-frame head concentrates on the current slice, and the window head focuses on a local spatio-temporal neighborhood around the query.}
    \label{fig:asr_frame_attention_application}
    \vspace{-2mm}
\end{figure*}

To further illustrate how the routed attention patterns are applied on real video frames, we visualize the attention distributions of three representative heads assigned to global, intra-frame, and window routing in Fig.~\ref{fig:asr_frame_attention_application}. For a fixed query token on the middle temporal slice, the global head distributes attention broadly across both space and time, the intra-frame head concentrates almost entirely on the current slice, and the window head focuses on a local spatio-temporal neighborhood centered around the query. This visualization provides an intuitive view of how different routed heads specialize to complementary receptive fields.


\begin{figure*}[t]
    \centering
    \includegraphics[width=0.9\textwidth]{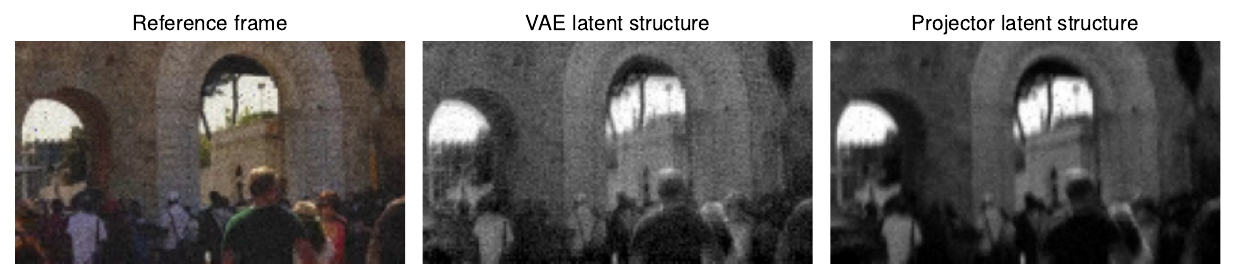}
    \vspace{-3mm}
    \caption{Visualization of encoder-free latent projection. Given the same LQ input, we compare latents produced by a frozen VAE encoder and by the proposed pixel-unshuffle plus linear projector. Both latents share the same shape. We visualize their channel-invariant spatial structures using the first principal component (PC1). The projector preserves scene structures highly consistent with the VAE latent, indicating that it serves as a lightweight alternative for VSR inference.}
    \label{fig:vae_projector_structure}
    \vspace{-2mm}
\end{figure*}

\subsubsection{Attention Specialization Routing}
The core idea of this routing strategy is to selectively replace global attention heads with localized alternatives, based on the similarities of their attention heatmaps. To this end, we first present the detailed formulations of two attention mechanisms, intra-frame and window attention, which we identify as the key localized patterns. 

\textbf{Intra-Frame Attention.} Let the latents of the input LQ video be denoted as $\mathbf{z}_{\text{L}}\in\mathbb{R}^{C\times T\times H\times W}$, where $C$ is the latent dimension, $T$ is the temporal length (number of frames), and $H$, $W$ are the spatial height and width, respectively. In diffusion transformers (DiTs), the query, key, and value tokens are obtained by applying separate learned linear projections to $\mathbf{z}_{\text{L}}$. For each query token $q_{t,h,w}\in \mathbb{R}^d$ extracted from spatial position $(h,w)$ in frame $t$, where $d$ denotes the feature dimension, the key and value sets are constructed by gathering all tokens within the same frame $t$. The intra-frame attention output is thus computed as:
\begin{align}
\mathrm{Attn}_{\text{intra}}
=
\frac{\sum_{h',w'}
\exp\!\left(\frac{\mathbf{q}_{t,h,w}^{\top}\mathbf{k}_{t,h',w'}}{\sqrt{d}}\right)
\mathbf{v}_{t,h',w'}
}{
\sum_{h'',w''}
\exp\!\left(\frac{\mathbf{q}_{t,h,w}^{\top}\mathbf{k}_{t,h'',w''}}{ \sqrt{d}}\right)
}.
\end{align}

\textbf{Window Attention.} For each query token $\mathbf{q}_{t,h,w} \in \mathbb{R}^d$ at position $(t,h,w)$, the attention is restricted to a local spatiotemporal neighborhood. Specifically, we define a window of size $(P_t, P_h, P_w)$ centered at $(t,h,w)$, and gather all tokens whose indices fall within this window:

\begin{equation}
\mathcal{N}(t,h,w)=
\left\{(t',h',w')\;\middle|\;
\begin{array}{l}
|t'-t| \le P_t/2,\\
|h'-h| \le P_h/2,\\
|w'-w| \le P_w/2
\end{array}
\right\}.
\end{equation}

The window attention output is then computed as:
\begin{equation}
\mathrm{Attn}_{\text{win}}
= 
\frac{\sum_{\mathcal{N}(t,h,w)} \exp\big(\frac{\mathbf{q}_{t,h,w}^\top \mathbf{k}_{t',h',w'}}{\sqrt{d}}\big)\mathbf{v}_{t',h',w'}}
{\sum_{\mathcal{N}(t,h,w)} \exp\big(\frac{\mathbf{q}_{t,h,w}^\top \mathbf{k}_{t'',h'',w''}}{\sqrt{d}}\big)}.
\end{equation}
To better exploit head-level specialization, we propose a simple algorithm to perform the routing. As illustrated in Algorithm~\ref{alg:routing}, given a target ratio $\rho$ that specifies the proportion of attention heads to be retained as global, we evaluate each head on a set of videos by measuring the KL-divergence between its global attention heatmap and those from intra-frame and window attention. The smaller divergence value is taken as the head score $s_h$. Heads are then ranked by $s_h$ and sequentially replaced with the corresponding localized alternative until the remaining global heads meet the ratio $\rho$.
\begin{algorithm}[h]
\caption{Attention Specialization Routing}
\label{alg:routing}
\begin{algorithmic}[1]
\State \textbf{Input:} DiT $\mathcal{DN}_\theta$, global-head ratio $\rho\in[0,1]$
\State \textbf{Output:} Assignment map $\mathcal{A}$ (head $\rightarrow$ \{global, intra-frame, window\})
\State Initialize $\mathcal{A}(h)\leftarrow\text{global}$ for all heads;\quad $N\leftarrow$ number of heads in all layers
\State \textbf{For each head $h$ across all layers:}
\State \hspace{1em} obtain attention map $M^g_{h},\,M^i_{h},\,M^w_{h}$ with global, intra-frame, and window attention
\State \hspace{1em} $s^i_{h}\leftarrow\mathbb{E}\left[\mathrm{KL}(M^g_{h}\,\|\,M^i_{h})\right]$;\quad
                     $s^w_{h}\leftarrow\mathbb{E}\left[\mathrm{KL}(M^g_{h}\,\|\,M^w_{h})\right]$
\State \hspace{1em} $s_{h}\leftarrow\min\{s^i_{h},s^w_{h}\}$;\quad
                     $m_{h}\leftarrow\arg\min_{m\in\{i, w\}} s^m_{h}$
\State Sort all heads by $s_{h}$ (ascending);\quad $K\leftarrow\lceil \rho N\rceil$
\State Set $\mathcal{A}(h_j)\leftarrow m_{h_j}$ for $j=1,\dots,N-K$
\State \Return $\mathcal{A}$
\end{algorithmic}
\end{algorithm}

\subsubsection{Redundancy Beyond Attention}
For video super-resolution, diffusion models typically incorporate a VAE encoder to map frames into the latent space and a prompt extractor for conditioning~\citep{zhou2024upscale}. In our setting, these components introduce additional computational overhead. Therefore, we replace the VAE encoder with an extremely lightweight pixel-unshuffle~\citep{chen2025adversarial} operation followed by a linear projection. Moreover, since the prompt extractor is derived from the same LQ video without introducing additional information, we omit it as well and use an empty prompt for conditioning.

To further examine whether the expensive VAE encoder is necessary for VSR inference, we visualize the latent representations produced by two alternative encoding paths given the same LQ input. The first path follows the conventional design, where the LQ video is bicubic-upsampled and encoded by a frozen VAE encoder. The second path replaces this process with pixel-unshuffle followed by the learned linear projector. As shown in Fig.~\ref{fig:vae_projector_structure}, although the two latents are not constrained to share an identical channel basis, their channel-invariant spatial structures remain highly consistent. In particular, the projector latent preserves the main scene layout and object boundaries, such as the arch, windows, and foreground people. This observation suggests that the learned projector provides a structurally meaningful latent representation that is well aligned with that of the VAE encoder, supporting the removal of the expensive VAE encoder during VSR inference.

\begin{table*}[t]
    \footnotesize
    \hspace{-2.mm}
    \centering
    \caption{Quantitative comparison on synthetic and real-world datasets. The best and second best results are colored with \textcolor{red}{red} and \textcolor{blue}{blue}. OASIS excels across multiple datasets and metrics.}
    \vspace{-2.5mm}
    \label{tab:quantitative}
    \resizebox{\linewidth}{!}{
        \setlength{\tabcolsep}{1.8mm}
        \begin{tabular}{l|l|c|c|c|c|c|c|c|c}
            \toprule[0.1em]
            \rowcolor{tablecolor}
             & & RealBasicVSR & Upscale-A-Video & MGLD-VSR & VEnhancer & STAR & SeedVR & SeedVR2 & OASIS (ours) \\
            \rowcolor{tablecolor} \multirow{-2}{*}{Dataset} & \multirow{-2}{*}{Metric} & CVPR 2022 & CVPR 2024 & ECCV 2024 & arXiv 2024 & ICCV 2025 & ICCV 2025 & arXiv 2025 & 2025 \\
            \midrule[0.1em]
            \multirow{6}{*}{UDM10}
                & PSNR $\uparrow$         & 24.13 & 21.72 & 24.23 & 21.32 & 23.47 & 24.39 & \textcolor{blue}{25.39} & \textcolor{red}{25.63} \\
                & SSIM $\uparrow$         & 0.6801 & 0.5913 & 0.6957 & 0.6811 & 0.6804 & 0.7083 & \textcolor{blue}{0.7564} & \textcolor{red}{0.7579} \\
                & LPIPS $\downarrow$      & 0.3908 & 0.4116 & 0.3272 & 0.4344 & 0.4242 & 0.3417 & \textcolor{blue}{0.2868} & \textcolor{red}{0.2452} \\
                & CLIP-IQA $\uparrow$     & 0.3494 & \textcolor{blue}{0.4697} & 0.4557 & 0.2852 & 0.2417 & 0.2869 & 0.2906 & \textcolor{red}{0.5510} \\
                & DOVER $\uparrow$        & \textcolor{blue}{0.7564} & 0.7291 & 0.7264 & 0.4576 & 0.4830 & 0.5493 & 0.5646 & \textcolor{red}{0.7863} \\
                & $E^*_{warp} \downarrow$ & 3.10   & 3.97   & 3.59   & \textcolor{red}{1.03} & 2.08   & 3.84   & 2.59   & \textcolor{blue}{1.94} \\
            \midrule
            \multirow{6}{*}{SPMCS}
                & PSNR $\uparrow$         & 22.17 & 18.81 & \textcolor{blue}{22.39} & 18.58 & 21.24 & 21.73 & 22.36 & \textcolor{red}{22.75} \\
                & SSIM $\uparrow$         & 0.5638 & 0.4113 & 0.5896 & 0.4850 & 0.5441 & 0.5803 & \textcolor{red}{0.6136} & \textcolor{blue}{0.5904} \\
                & LPIPS $\downarrow$      & 0.3662 & 0.4468 & 0.3263 & 0.5358 & 0.5257 & 0.3297 & \textcolor{blue}{0.2905} & \textcolor{red}{0.2634} \\
                & CLIP-IQA $\uparrow$     & 0.3513 & \textcolor{red}{0.5248} & 0.4348 & 0.3188 & 0.2646 & 0.3946 & 0.4086 & \textcolor{blue}{0.4693} \\
                & DOVER $\uparrow$        & 0.6753 & \textcolor{blue}{0.7171} & 0.6754 & 0.4284 & 0.3204 & 0.6150 & 0.6251 & \textcolor{red}{0.7242} \\
                & $E^*_{warp} \downarrow$ & 1.88   & 4.22   & 1.68   & 1.19   & \textcolor{red}{1.01} & 1.83   & 1.24   & \textcolor{blue}{1.15} \\
            \midrule
            \multirow{6}{*}{YouHQ40}
                & PSNR $\uparrow$         & 22.39 & 19.62 & \textcolor{blue}{23.17} & 19.78 & 22.43 & 21.96 & 23.61 & \textcolor{red}{23.75} \\
                & SSIM $\uparrow$         & 0.5895 & 0.4824 & 0.6194 & 0.5911 & 0.6276 & 0.5920 & \textcolor{red}{0.6771} & \textcolor{blue}{0.6417} \\
                & LPIPS $\downarrow$      & 0.4091 & 0.4268 & 0.3608 & 0.4742 & 0.4744 & 0.3466 & \textcolor{blue}{0.2754} & \textcolor{red}{0.2608} \\
                & CLIP-IQA $\uparrow$     & 0.3964 & \textcolor{red}{0.5258} & 0.4657 & 0.3309 & 0.2805 & 0.4123 & 0.3811 & \textcolor{blue}{0.4817} \\
                & DOVER $\uparrow$        & 0.8596 & 0.8596 & 0.8446 & 0.6957 & 0.5525 & \textcolor{blue}{0.8618} & 0.8384 & \textcolor{red}{0.8700} \\
                & $E^*_{warp} \downarrow$ & 3.08   & 6.84   & 3.45   & \textcolor{red}{0.95} & 3.39   & 3.44   & 3.42   & \textcolor{blue}{2.74} \\
            \midrule\midrule
            \multirow{6}{*}{RealVSR}
                & PSNR $\uparrow$         & \textcolor{blue}{22.00} & 20.74 & \textcolor{red}{22.08} & 15.75 & 17.43 & 20.44 & 20.20 & 21.14 \\
                & SSIM $\uparrow$         & \textcolor{red}{0.7166} & 0.5681 & 0.6805 & 0.4002 & 0.5215 & 0.6792 & \textcolor{blue}{0.6977} & 0.6212 \\
                & LPIPS $\downarrow$      & \textcolor{blue}{0.2036} & 0.4163 & 0.2241 & 0.3784 & 0.2943 & 0.2416 & 0.2197 & \textcolor{red}{0.2018} \\
                & CLIP-IQA $\uparrow$     & 0.3538 & 0.2134 & \textcolor{blue}{0.4109} & 0.3880 & 0.3641 & 0.2924 & 0.2887 & \textcolor{red}{0.4357} \\
                & DOVER $\uparrow$        & 0.7384 & 0.3587 & 0.7354 & \textcolor{blue}{0.7637} & 0.7051 & 0.6747 & 0.7209 & \textcolor{red}{0.7800} \\
                & $E^*_{warp} \downarrow$ & 4.72   & \textcolor{red}{1.00} & 3.03   & 5.15   & 9.88   & 3.62   & 4.77   & \textcolor{blue}{2.63} \\
            \midrule
            \multirow{6}{*}{MVSR4x}
                & PSNR $\uparrow$         & 21.80 & 22.35 & 22.58 & 20.50 & 22.42 & 22.16 & 21.72 & \textcolor{red}{22.66} \\
                & SSIM $\uparrow$         & 0.7045 & 0.7327 & 0.7399 & 0.7117 & 0.7421 & 0.7407 & \textcolor{red}{0.7566} & \textcolor{blue}{0.7428} \\
                & LPIPS $\downarrow$      & 0.4235 & 0.4012 & 0.3486 & 0.4471 & 0.4311 & 0.4543 & 0.3667 & \textcolor{red}{0.3246} \\
                & CLIP-IQA $\uparrow$     & 0.4118 & 0.3235 & 0.3738 & 0.3104 & 0.2674 & 0.2271 & 0.2243 & \textcolor{red}{0.5711} \\
                & DOVER $\uparrow$        & \textcolor{blue}{0.6846} & 0.4276 & 0.6062 & 0.3164 & 0.2137 & 0.1554 & 0.2219 & \textcolor{red}{0.7243} \\
                & $E^*_{warp} \downarrow$ & 1.69   & \textcolor{blue}{0.66} & 1.51   & \textcolor{blue}{0.62} & \textcolor{red}{0.61} & 2.28   & 1.33   & 0.87 \\
            \bottomrule[0.1em]
        \end{tabular}
    }
    \vspace{-2mm}
\end{table*}

\subsection{Progressive Training}
\subsubsection{Two-Stage Curriculum Learning}
After reducing structural redundancy, the remaining challenge is optimization under complex real-world degradations. Real-world videos often exhibit temporal inconsistency: unstable camera motion, compression, or blur may affect different frames with different severity. Directly training on such inconsistent degradations increases the learning burden, especially for a one-step restoration model. We therefore adopt a progressive curriculum that moves from easier temporally consistent degradations to more realistic inconsistent degradations.

The progressive training consists of two stages. In the first stage, we employ the second-order degradation model~\citep{wang2021real} to generate synthetic training data. Specifically, we randomly apply Gaussian noise, blur, and compression artifacts (image and video) to HQ videos. To control the difficulty, all frames within a video are assigned the same type and severity of degradation, ensuring temporal consistency across the sequence.

In the second stage, degradations are made temporally inconsistent across frames. To avoid excessive discontinuities, degradations are generated sequentially across frames~\citep{chan2022investigating}, with the type and severity conditioned on the previous frame and stochastically perturbed with a predefined probability. This setting more faithfully reflects real-world scenarios and further improves the model’s robustness to diverse frame-wise variations. Our ablation studies demonstrate that progressive training clearly outperforms direct training on temporally inconsistent degradations.

\subsubsection{Training Objectives}
In our progressive training, the two stages share the same training objective.  Apart from the latent reconstruction loss for the one-step diffusion model, we further introduce perceptual and temporal losses in pixel space to enhance both visual fidelity and temporal consistency.

\textbf{Latent Reconstruction Loss.} Unlike standard diffusion models that optimize a noise-prediction loss~\citep{ho2020denoising}, OASIS employs a latent reconstruction objective, which is defined as an MSE loss between $\hat{\mathbf{z}}_{\text{H}}$ and its ground-truth counterpart $\mathbf{z}_{\text{H}}$ over a mini-batch of size $B$:
\begin{equation}
    \mathcal{L}_\text{latent}\left(\hat{\mathbf{z}}_{\text{H}}, \mathbf{z}_{\text{H}} \right)=\mathcal{L}_2\left(\hat{\mathbf{z}}_{\text{H}}, \mathbf{z}_{\text{H}} \right)=\frac{1}{B}\sum_{i=1}^B\|\hat{\mathbf{z}}_{\text{H}}- \mathbf{z}_{\text{H}}\|^2.
\end{equation}

\begin{figure*}[t]
\scriptsize
\centering
\begin{tabular}{cc}
\resizebox{1.\textwidth}{!}{
    \begin{tabular}{c}
    \hspace{-4mm}
    \includegraphics[width=0.174\textwidth, height=0.166\textwidth]{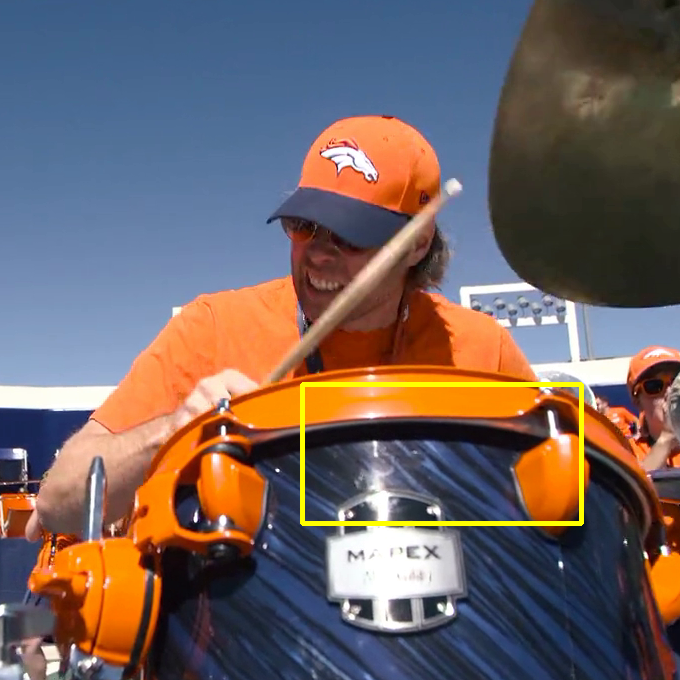}
    \\
    UDM10: 003
    \end{tabular}
    \hspace{-4.5mm}
    \begin{tabular}{cccc}
    \includegraphics[width=0.14\textwidth]{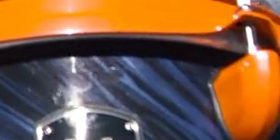} \hspace{-4mm} &
    \includegraphics[width=0.14\textwidth]{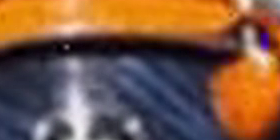} \hspace{-4mm} &
    \includegraphics[width=0.14\textwidth]{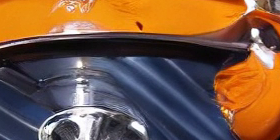} \hspace{-4mm} &
    \includegraphics[width=0.14\textwidth]{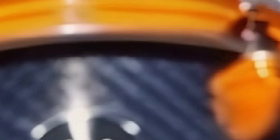} 
    \\
    GT \hspace{-4mm} &
    Bicubic \hspace{-4mm} &
    Upscale-A-Video \hspace{-4mm} &
    VEnhancer \\
    \includegraphics[width=0.14\textwidth]{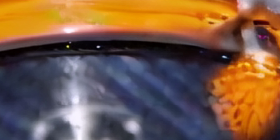} \hspace{-4mm} &
    \includegraphics[width=0.14\textwidth]{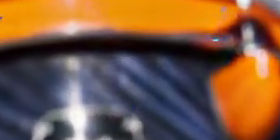} \hspace{-4mm} &
    \includegraphics[width=0.14\textwidth]{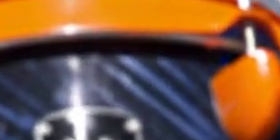} \hspace{-4mm} &
    \includegraphics[width=0.14\textwidth]{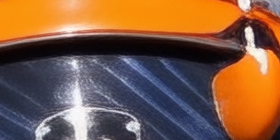}  
    \\ 
    STAR \hspace{-4mm} &
    SeedVR \hspace{-4mm} &
    SeedVR2 \hspace{-4mm} &
    OASIS (ours)
    \\
    \end{tabular}
}
\\

\resizebox{1.\textwidth}{!}{
    \begin{tabular}{c}
    \hspace{-4mm}
    \includegraphics[width=0.174\textwidth, height=0.166\textwidth]{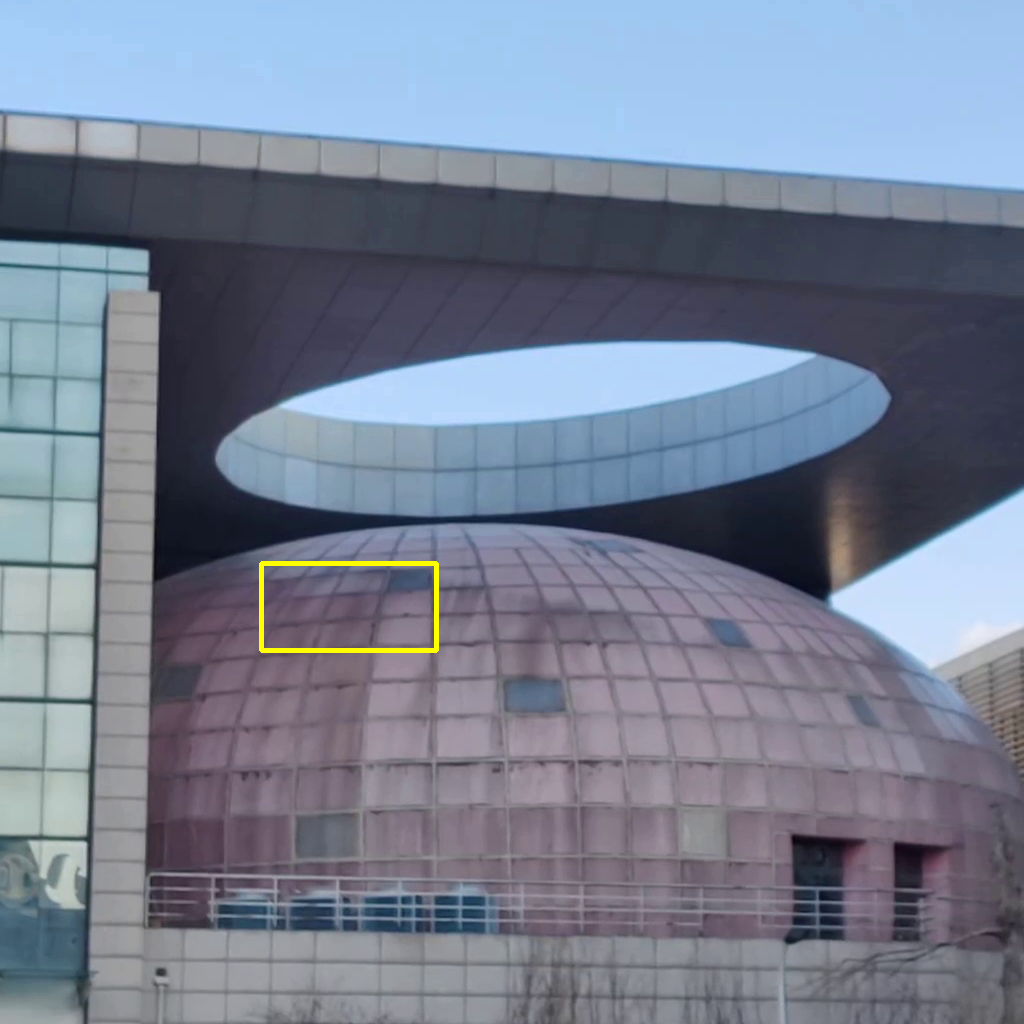}
    \\
    MVSR4x: 013
    \end{tabular}
    \hspace{-4.5mm}
    \begin{tabular}{cccc}
    \includegraphics[width=0.14\textwidth]{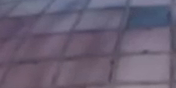} \hspace{-4mm} &
    \includegraphics[width=0.14\textwidth]{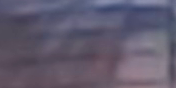} \hspace{-4mm} &
    \includegraphics[width=0.14\textwidth]{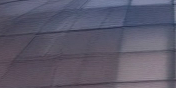} \hspace{-4mm} &
    \includegraphics[width=0.14\textwidth]{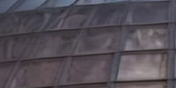} 
    \\
    GT \hspace{-4mm} &
    Bicubic \hspace{-4mm} &
    Upscale-A-Video \hspace{-4mm} &
    VEnhancer \\
    \includegraphics[width=0.14\textwidth]{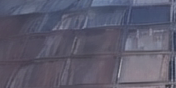} \hspace{-4mm} &
    \includegraphics[width=0.14\textwidth]{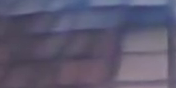} \hspace{-4mm} &
    \includegraphics[width=0.14\textwidth]{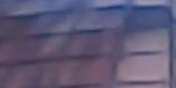} \hspace{-4mm} &
    \includegraphics[width=0.14\textwidth]{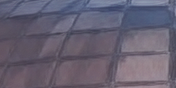}  
    \\ 
    STAR \hspace{-4mm} &
    SeedVR \hspace{-4mm} &
    SeedVR2 \hspace{-4mm} &
    OASIS (ours)
    \\
    \end{tabular}
}
\\

\end{tabular}
\vspace{-2mm}
\caption{Visual comparisons on synthetic and real-world datasets for $\times$4 VSR. OASIS yields clean reconstructions while preserving contours and fine-scale surface patterns.}
\label{fig:main_visual}
\end{figure*}

\begin{figure*}[t]
    \centering

    \begin{minipage}[t]{0.46\linewidth}
        \vspace{0pt}
        \centering
        \includegraphics[width=\linewidth]{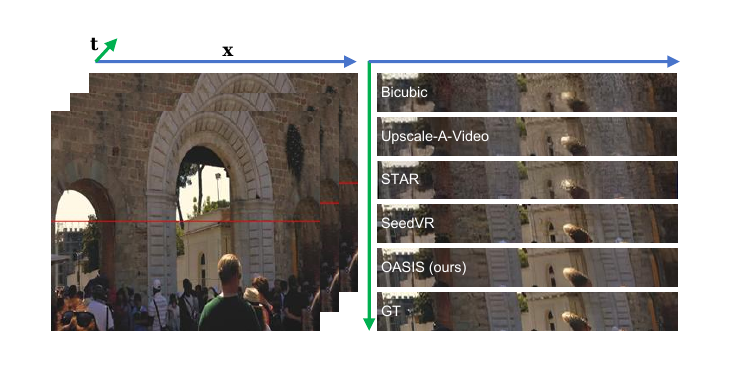}
        \vspace{-7mm}
        \captionof{figure}{Comparison of temporal consistency. The temporal profile is obtained by stacking the \textcolor{red}{red} line across frames. Our method produces smoother frame transitions that closely resemble the ground truth.}
        \label{fig:consist}
    \end{minipage}
    \hfill
    \begin{minipage}[t]{0.51\linewidth}
        \vspace{0pt}
        \centering
        \scriptsize
        \captionof{table}{Comparison of inference steps (Step), number of parameters (Params), running time (Time), and multiply accumulate operations (MACs) of different diffusion-based VSR methods on a 33-frame 720$\times$1280 video. }
        \label{tab:time}
        \vspace{-1mm}
        \resizebox{\linewidth}{!}{
        \begin{tabular}{l|cccc}
            \toprule[0.1em]
            \rowcolor{tablecolor}
            Method & Step & Params (B) & Time (s) & MACs (T) \\
            \midrule[0.1em]
            Upscale-A-Video & 30 & \textcolor{red}{1.09} & 283.70 & 9,084.73 \\
            MGLD-VSR & 50 & 1.57 & 429.48 & 8,528.70 \\
            VEnhancer & 15 & 2.50 & 122.48 & 3,056.16 \\
            STAR & 15 & 2.49 & 176.53 & 4,281.67 \\
            SeedVR & 50 & 3.40 & 207.13 & 8,243.13 \\
            SeedVR2 & \textcolor{red}{1} & 3.40 & 30.81 & 86.97 \\
            OASIS (ours) & \textcolor{red}{1} & 1.42 & \textcolor{red}{4.97} & \textcolor{red}{51.43} \\
            \bottomrule[0.1em]
        \end{tabular}
        }
    \end{minipage}
    \vspace{-2mm}
\end{figure*}

\begin{figure*}[t]
    \centering
    \begin{adjustwidth}{2mm}{-2mm}
        \centering
        \resizebox{1.\textwidth}{!}{
        \begin{tabular}{ccccccc} 
            \includegraphics[width=0.265\linewidth]{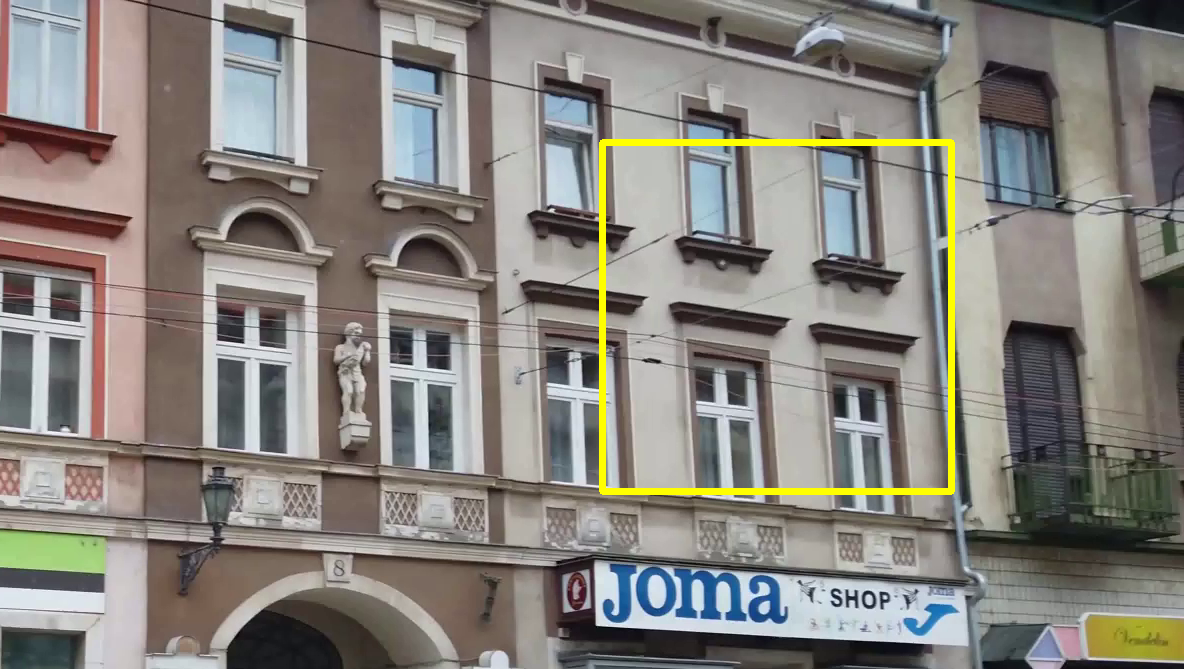} \hspace{-4mm} &
            \includegraphics[width=0.150\linewidth]{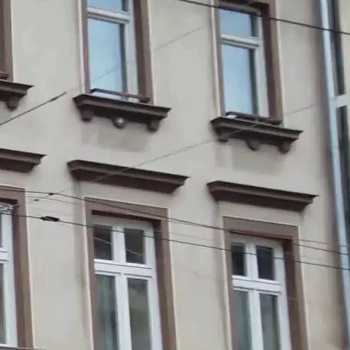} \hspace{-4mm} &
            \includegraphics[width=0.150\linewidth]{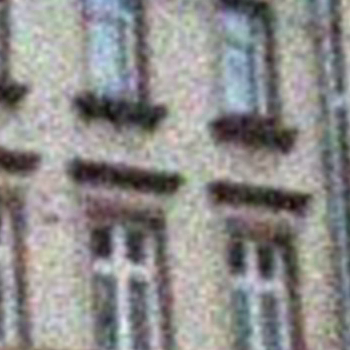} \hspace{-4mm} &
            \includegraphics[width=0.150\linewidth]{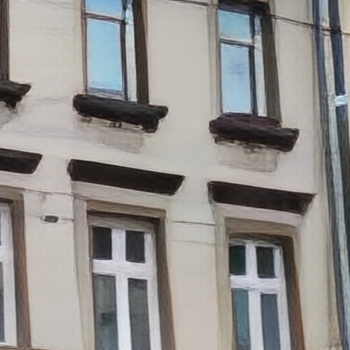} \hspace{-4mm} &
            \includegraphics[width=0.150\linewidth]{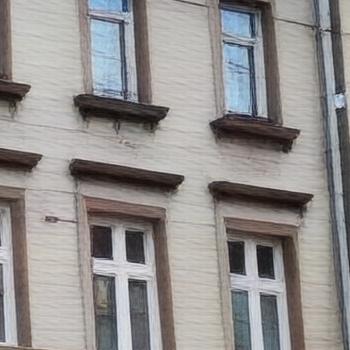} \hspace{-4mm} &
            \includegraphics[width=0.150\linewidth]{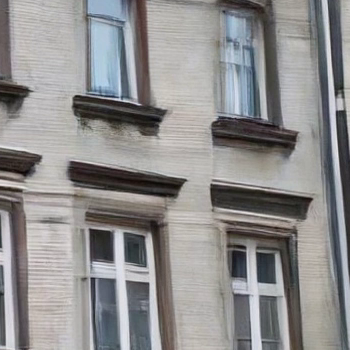} \hspace{-4mm} &
            \includegraphics[width=0.150\linewidth]{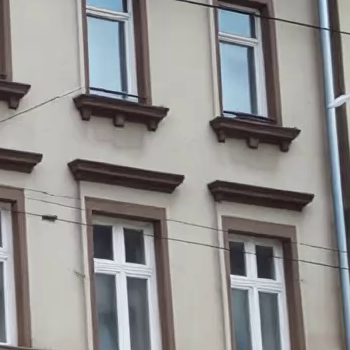} \\
            YouHQ40: 030 \hspace{-4mm} & GT \hspace{-4mm} & Bicubic \hspace{-4mm} & Global \hspace{-4mm} & Intra-Frame \hspace{-4mm} & Window \hspace{-4mm} & ASR (ours) \\
        \end{tabular}
        }\\
        \resizebox{1.\textwidth}{!}{
        \begin{tabular}{ccccccc} 
            \includegraphics[width=0.265\linewidth]{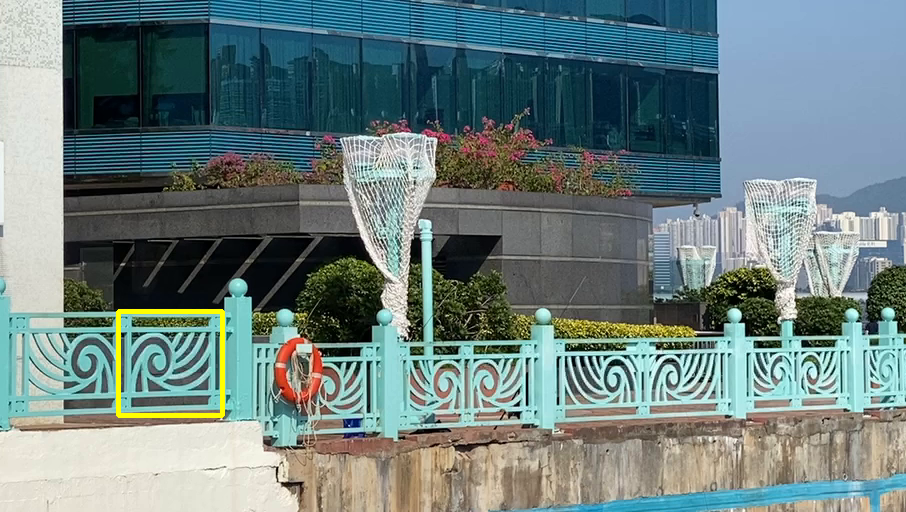} \hspace{-4mm} &
            \includegraphics[width=0.150\linewidth]{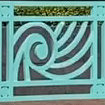} \hspace{-4mm} &
            \includegraphics[width=0.150\linewidth]{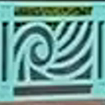} \hspace{-4mm} &
            \includegraphics[width=0.150\linewidth]{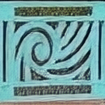} \hspace{-4mm} &
            \includegraphics[width=0.150\linewidth]{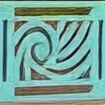} \hspace{-4mm} &
            \includegraphics[width=0.150\linewidth]{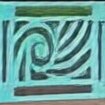} \hspace{-4mm} &
            \includegraphics[width=0.150\linewidth]{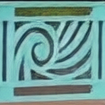} \\
            RealVSR: 317 \hspace{-4mm} & GT \hspace{-4mm} & Bicubic \hspace{-4mm} & Global \hspace{-4mm} & Intra-Frame \hspace{-4mm} & Window \hspace{-4mm} & ASR (ours) \\
        \end{tabular}
        }\\
    \end{adjustwidth}
    \refstepcounter{figure}
    \label{fig:ablation}
    \begin{center}
    \small Fig.~\thefigure. Visual comparisons between ASR and different attention patterns.
    \end{center}
\end{figure*}

\begin{table*}[t]
\centering
\vspace{-4mm}
\caption{Quantitative results of the ablation study on the MVSR4x dataset. Best results are in \textcolor{red}{red}.}
\vspace{-5mm}
\label{table:ablation}

\resizebox{\textwidth}{!}{
\subfloat[\footnotesize Ablation study of ASR.\label{abl:asr}]{
\vspace{-1mm}
\large
\setlength{\tabcolsep}{2mm}
\resizebox{0.47\linewidth}{!}{
\renewcommand{\arraystretch}{1.1}
\begin{tabular}{l|cccc|c}
\toprule[0.15em]
\rowcolor{tablecolor}
Method & PSNR$\uparrow$ & LPIPS$\downarrow$ & DOVER$\uparrow$ & $E^*_{warp}$$\downarrow$ & Time (s) \\
\midrule[0.15em]
Global & 22.35 & 0.3418 & 0.7098 & 1.13 & 10.53 \\
Intra-Frame & 22.10 & 0.3324 & 0.6730 & 1.75 & \textcolor{red}{2.38} \\
Window & 22.28 & 0.3408 & 0.6831 & 1.18 & 6.60 \\
ASR (ours) & \textcolor{red}{22.77} & \textcolor{red}{0.3242} & \textcolor{red}{0.7142} & \textcolor{red}{0.87} & 7.50 \\
\bottomrule[0.15em]
\end{tabular}
}}
\subfloat[\footnotesize Ablation study of redundancy beyond attention.\label{abl:beyond}]{
\vspace{-1mm}
\huge
\setlength{\tabcolsep}{2mm}
\resizebox{0.48\linewidth}{!}{
\renewcommand{\arraystretch}{1.29}
\begin{tabular}{l|cccc|cc}
\toprule[0.16em]
\rowcolor{tablecolor}
Method & PSNR$\uparrow$ & LPIPS$\downarrow$ & DOVER$\uparrow$ & $E^*_{warp}$$\downarrow$ & Time (s) & MACs (T) \\
\midrule[0.16em]
w/ VAE+Prompt & 22.33 & 0.3433 & 0.6994 & 1.21 & 7.76 & 57.55  \\
w/o VAE & 22.31 & \textcolor{red}{0.3407} & 0.7007 & 1.16 & 6.03 & 55.79 \\
w/o Prompt & \textcolor{red}{22.41} & 0.3424 & 0.6901 & 1.17 & 6.84 & 53.09 \\
w/o VAE+Prompt & 22.35 & 0.3418 & \textcolor{red}{0.7098} & \textcolor{red}{1.13} & \textcolor{red}{4.97} & \textcolor{red}{51.43} \\
\bottomrule[0.16em]
\end{tabular}
}}
}
\\
\vspace{1mm}
\resizebox{\textwidth}{!}{
\subfloat[\footnotesize Ablation study of progressive training.\label{abl:training}]{
\vspace{-1mm}
\large
\setlength{\tabcolsep}{2mm}
\resizebox{0.44\textwidth}{!}{
\renewcommand{\arraystretch}{1.2}
\begin{tabular}{l|ccccc}
\toprule[0.15em]
\rowcolor{tablecolor}
Method & PSNR$\uparrow$ & SSIM$\uparrow$ & LPIPS$\downarrow$ & DOVER$\uparrow$ & $E^*_{warp}$$\downarrow$ \\
\midrule[0.15em]
S1 & 22.23 & 0.7369 & 0.3378 & 0.6978 & 1.31 \\
S2 & 22.35 & 0.7336 & 0.3418 & 0.7098 & 1.13 \\
S1+S2 (ours) & \textcolor{red}{22.64} & \textcolor{red}{0.7432} & \textcolor{red}{0.3258} & \textcolor{red}{0.7144} & \textcolor{red}{0.97} \\
\bottomrule[0.15em]
\end{tabular}
}}
\subfloat[\footnotesize Ablation study of training loss functions.\label{abl:loss}]{
\vspace{-1mm}
\small
\setlength{\tabcolsep}{2mm}
\resizebox{0.5\textwidth}{!}{
\renewcommand{\arraystretch}{1.17}
\begin{tabular}{ccc|ccccc}
\toprule[0.15em]
\rowcolor{tablecolor}
$\mathcal{L}_\text{latent}$ & $\mathcal{L}_{\text{per}}$ & $\mathcal{L}_{\text{warp}}$ & PSNR$\uparrow$ & SSIM$\uparrow$ & LPIPS$\downarrow$ & DOVER$\uparrow$ & $E^*_{warp}$$\downarrow$ \\
\midrule[0.15em]
\checkmark & & & 21.84 & 0.7744 & 0.4035 & 0.5871 & 2.07 \\
\checkmark & \checkmark & & 22.28 & \textcolor{red}{0.7351} & 0.3432 & 0.6896 & 1.54 \\
\checkmark & \checkmark & \checkmark & \textcolor{red}{22.35} & 0.7336 & \textcolor{red}{0.3418} & \textcolor{red}{0.7098} & \textcolor{red}{1.13} \\
\bottomrule[0.15em]
\end{tabular}
}}
}
\vspace{-5mm}
\end{table*}

\textbf{Perceptual Loss.} Although the latent reconstruction loss provides direct supervision to the DiT, the target latent $\mathbf{z}_{\text{H}}$ obtained from the VAE encoder typically deviates slightly from the true HQ latent representations. To remedy this, we introduce a perceptual loss in pixel space, combining MSE and LPIPS~\citep{zhang2018unreasonable} to balance accuracy and visual quality:
\begin{equation}
    \mathcal{L}_{\text{per}}(\hat{\mathbf{V}}_{\text{H}}, \mathbf{V}_{\text{H}} )=\mathcal{L}_2(\hat{\mathbf{V}}_{\text{H}}, \mathbf{V}_{\text{H}} )+\mathcal{L}_{\text{LPIPS}}(\hat{\mathbf{V}}_{\text{H}}, \mathbf{V}_{\text{H}} ).
\end{equation}
\textbf{Temporal Loss.} Supervision in pixel space is frame-wise and lacks explicit enforcement of temporal consistency. To strengthen the coherence of the restored HQ video, we extract optical flow~\citep{teed2020raft} from the ground-truth video and warp each predicted frame toward its neighboring frame. The temporal loss is defined as the MAE between the warped frame and its neighbor:
\begin{equation}
\begin{aligned}
\mathcal{L}_{\text{warp}}
= \sum_{i=1}^M\Big(
& \|\operatorname{Warp}(\hat{\mathbf{V}}_{\text{H}}^i, \mathbf{O}^{\text{bw}, i}_{\text{GT}})
   - \hat{\mathbf{V}}_{\text{H}}^{i+1} \|_1 \\
+{}&
  \|\operatorname{Warp}(\hat{\mathbf{V}}_{\text{H}}^i, \mathbf{O}^{\text{fw}, i}_{\text{GT}})
   - \hat{\mathbf{V}}_{\text{H}}^{i-1} \|_1
\Big).
\end{aligned}
\end{equation}
where $M$ is the number of frames, $\mathbf{O}^{\text{bw}, i}_{\text{GT}}$ and $\mathbf{O}^{\text{fw}, i}_{\text{GT}}$ are the backward and forward optical flow derived from the ground-truth video. The overall training objectives can thereby be expressed as:
\begin{equation}
\mathcal{L}=\mathcal{L}_\text{latent}+\mathcal{L}_{\text{per}}+\lambda_{\text{warp}}\cdot\mathcal{L}_{\text{warp}}.
\end{equation}

\section{Experiments}
\subsection{Experimental Settings}
\textbf{Datasets and Metrics.} We train our model on the HQ-VSR dataset~\citep{chen2025dove}, which contains 2,055 videos. To generate paired LQ-HQ data, we adopt the RealBasicVSR~\citep{chan2022investigating} degradation model and extend it to both temporally consistent and inconsistent settings. For evaluation, we follow prior works~\citep{zhou2024upscale, yang2024motion} on both synthetic datasets (UDM10~\citep{tao2017detail}, SPMCS~\citep{yi2019progressive}, YouHQ40~\citep{zhou2024upscale}) and real-world datasets (RealVSR~\citep{yang2021real}, MVSR4x~\citep{wang2023benchmark}). All experiments use a $\times$4 upscaling factor. We report reference metrics including PSNR, SSIM~\citep{wang2004image}, and LPIPS~\citep{zhang2018unreasonable}, together with no-reference and temporal metrics including CLIP-IQA~\citep{wang2023exploring}, DOVER~\citep{wu2023exploring}, and $E^*_{warp}$ (\ie, $E_{warp}$ ($\times 10^{-3}$)~\citep{lai2018learning}).

\textbf{Implementation Details.} Our method is built on Wan2.1~\citep{wan2025wan} (1.42B parameters). We train the DiT with our progressive strategy while freezing all other components. Training is conducted on 8 NVIDIA A6000 GPUs using AdamW~\citep{loshchilov2017decoupled} ($\beta_1$$=$0.9, $\beta_2$$=$0.999). Input videos consist of 17 frames at 320$\times$640 resolution, with a batch size of 8. OASIS is trained for 25,000 iterations per stage with a learning rate of 1$\times$10$^{-4}$. We set loss weight $\lambda_\text{warp}$ to 0.1, predefined timestep $T_\text{L}$ to 799, and window attention size to $(3, 5, 5)$. For ASR, $\rho$ is 0.4, with attention assignments derived from 50 training videos from the training set.

\subsection{Main Results}

\textbf{Quantitative Results.} As shown in Tab.~\ref{tab:quantitative}, OASIS achieves superior performance, achieving first or second place on 27 of 30 reported results. It delivers top scores in pixel-level (PSNR and SSIM) and perceptual (LPIPS) fidelity, maintains superiority on video quality metrics (CLIP-IQA and DOVER), and shows competitive temporal consistency ($E^*_{warp}$). These results highlight that OASIS provides the most outstanding and balanced overall performance.

\textbf{Qualitative Results.} Figure~\ref{fig:main_visual} compares OASIS with leading baselines on synthetic and real-world videos. While other methods can preserve coarse structures, they often suffer from oversmoothed textures, contour artifacts, blurred grids, or color shifts. In contrast, OASIS recovers realistic details while maintaining fidelity to the original video. We also provide the temporal consistency visualization in Fig.~\ref{fig:consist}, where our method delivers strong temporal coherence, yielding smooth frame-to-frame transitions while preserving accurate details.

\textbf{Running Time Comparisons.} We evaluate efficiency in terms of inference steps, number of parameters, running time, and multiply accumulate operations (MACs) in Tab.~\ref{tab:time}. All methods are evaluated on one NVIDIA A100-80G GPU, generating a 33-frame 720$\times$1280 video. Notably, OASIS achieves a significant reduction in computational cost, benefiting from its one-step diffusion design that accelerates inference, along with the attention specialization routine and the removal of the VAE encoder and prompt extractor that mitigate redundancy.

\subsection{Ablation Study}

This section studies the effect of each component. All models are trained on the HQ-VSR dataset~\citep{chen2025dove} with a batch size of 4. For standard training, where the progressive strategy is not applied, each model is trained for 30,000 iterations. Under the progressive strategy, training is split into two stages, with 15,000 iterations performed for each stage.

\textbf{Attention Specialization Routing (ASR).} We compare ASR against global, intra-frame, and window attention patterns. As shown in Fig.~\ref{fig:ablation}, the hybrid design of ASR restores textures and details more faithfully than the single-pattern alternatives. In Tab.~\ref{table:ablation}(a), ASR consistently outperforms other variants across metrics. Notably, ASR surpasses the global-only baseline despite having a smaller overall receptive field. This confirms that uniformly forcing all heads to search globally introduces unnecessary long-range interactions that can dilute local details, whereas ASR optimally aligns head patterns with their intrinsic behaviors. Moreover, compared with the original global-only design, ASR also provides higher efficiency, with the runtime measured on a 105-frame 720$\times$1280 video using only the DiT module. These results demonstrate the effectiveness of ASR in reducing redundancy while enhancing performance.

\textbf{Redundancy of VAE Encoder and Prompt Extractor.} We further study redundancy beyond attention, as shown in Tab.~\ref{table:ablation}(b). Following Upscale-A-Video, we adopt LLaVA~\citep{liu2023visual} as the prompt extractor. Removing either module improves efficiency with minor impact on performance, with runtime measured on a 33-frame 720×1280 video.

\begin{figure*}[t]
    \centering
    \includegraphics[width=0.9\textwidth]{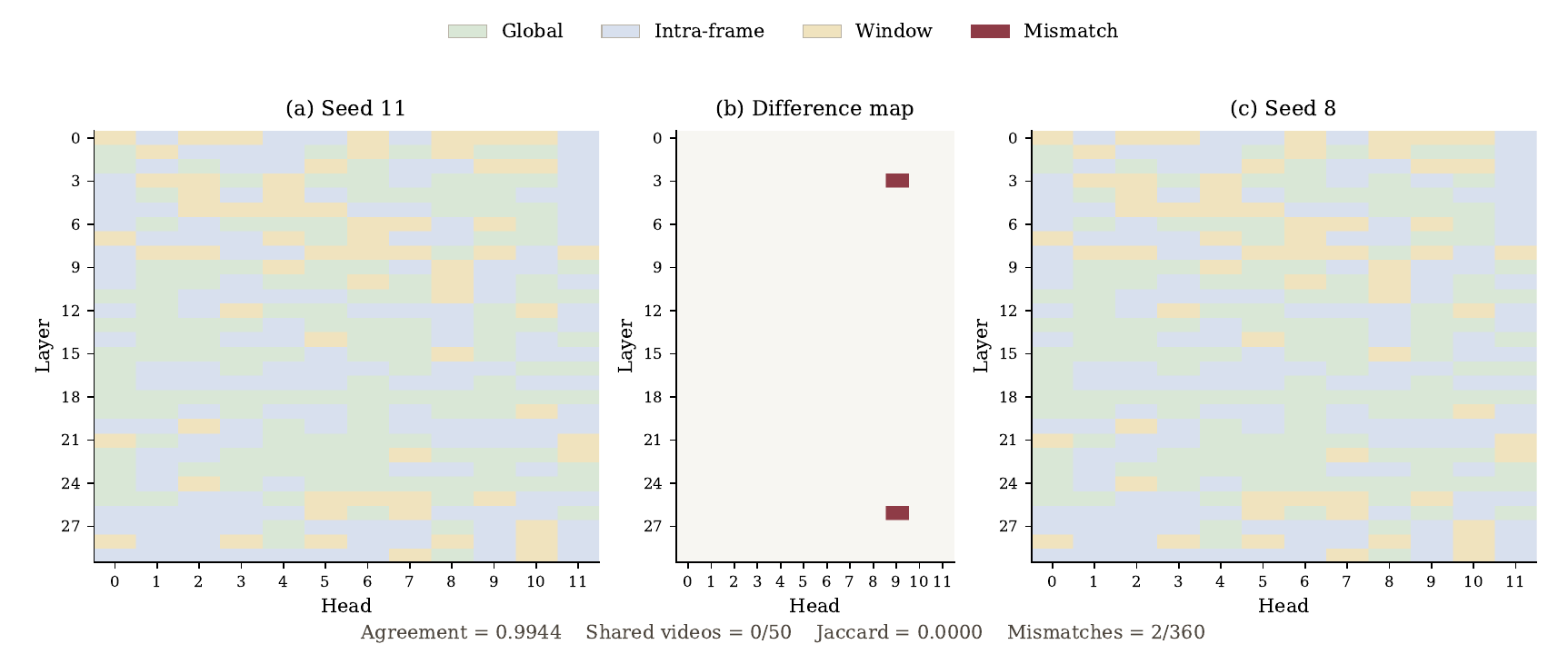}
    \vspace{-3mm}
    \caption{Stability of ASR routing across random 50-video subsets. We repeat the routing estimation procedure 20 times with different random seeds, each time sampling 50 training videos without replacement. Shown here is a zero-overlap pair, seeds 11 and 8. Although the two sampled subsets share no videos, their routing configurations differ at only 2 out of 360 layer-head positions, yielding 99.44\% agreement.}
    \label{fig:asr_stability}
    \vspace{-4mm}
\end{figure*}

\begin{figure}
\setlength{\tabcolsep}{0pt}
\begin{tabular}{@{}c@{}c@{}}
  \includegraphics[width=0.495\linewidth]{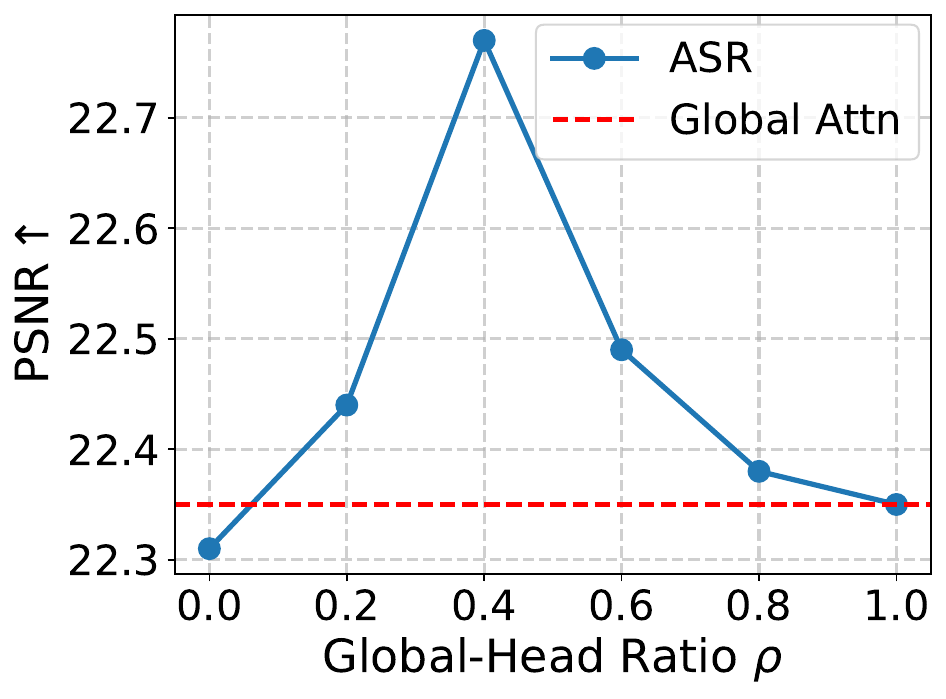} &
  \includegraphics[width=0.495\linewidth]{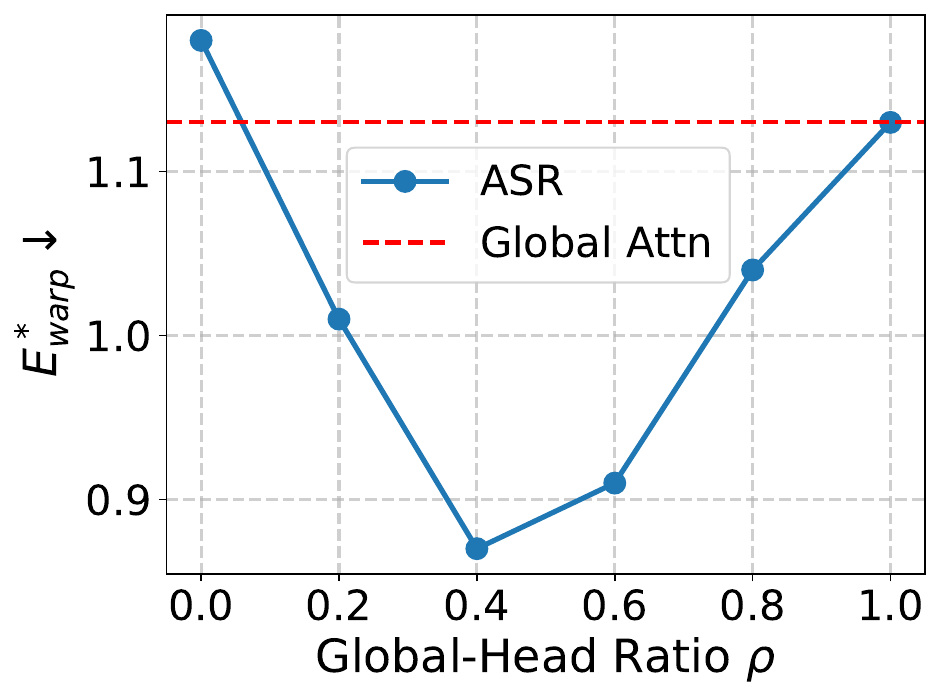}\\[-1mm]
  \scriptsize (a) PSNR vs.\ $\rho$ & \scriptsize (b) $E^*_{warp}$ vs.\ $\rho$
\end{tabular}
\vspace{-2mm}
\caption{Effect of global-head ratio $\rho$ on PSNR and $E^*_{warp}$ metrics. The results are evaluated on the MVSR4x dataset. Global Attn refers to the baseline using the global attention only.}
\label{fig:rho}
\vspace{-5mm}
\end{figure}

\textbf{Global-Head Ratio ($\rho$).} We evaluate ASR under different global-head ratios $\rho$ in Fig.~\ref{fig:rho}. At $\rho$$=$0.4, the model achieves the highest fidelity and temporal consistency, corresponding to the most effective specialization assignment. Notably, most values of $\rho$ outperform the global-only baseline, underscoring the soundness of the proposed attention specialization routing.

\textbf{Stability of ASR.} A potential concern is that the routing configuration may depend on the particular 50 training videos used for estimation. To evaluate this sensitivity, we repeat the routing estimation procedure 20 times using different random seeds, each time sampling 50 training videos without replacement and deriving one routing configuration. The resulting ASR patterns are highly consistent across runs, with a mean pairwise configuration agreement of 96.81\%, while the mean Jaccard overlap between sampled video subsets is only 0.0131. As a concrete example, Fig.~\ref{fig:asr_stability} visualizes one zero-overlap pair, seeds 11 and 8, whose routing configurations differ at only 2 out of 360 layer-head positions, yielding 99.44\% agreement. These results indicate that ASR captures stable head-level tendencies rather than overfitting to a particular 50-video subset.

\textbf{Progressive Training Strategy.} We compare standard training against our progressive training in Tab.~\ref{table:ablation}(c). Training with stage 1 (S1) alone results in poor temporal consistency, while training with stage 2 (S2) alone also leads to suboptimal performance. In contrast, the progressive strategy (S1+S2) delivers clear improvements under the same number of training iterations. This highlights the effectiveness of progressive training in guiding the model from simple to complex degradations and enhancing robustness in real-world scenarios.

\textbf{OASIS Training Loss Functions.} As shown in Tab.~\ref{table:ablation}(d), using only the latent reconstruction loss results in poor performance, indicating that latent supervision alone is insufficient. In contrast, introducing the perceptual loss yields substantial improvements on all metrics, demonstrating the critical role of pixel-level supervision. Finally, adding the temporal loss further improves the $E^*_{warp}$ score, highlighting its importance in enhancing temporal consistency.

\section{Conclusion}
We propose OASIS, an efficient one-step diffusion model for real-world VSR. Recognizing the task mismatch between video generation and conditioned restoration, OASIS effectively mitigates the redundancy inherent in pretrained diffusion models. Specifically, it incorporates an attention specialization routing that assigns attention heads to global or localized patterns according to their intrinsic behaviors, coupled with the elimination of redundant input conditioning modules such as the VAE encoder and prompt extractor. This task-specific adaptation significantly reduces computational overhead while preserving useful pretrained knowledge. Moreover, to ease the learning burden caused by these architectural simplifications, we design a progressive training strategy that guides the model from simple to complex degradations. Extensive experiments demonstrate that OASIS achieves state-of-the-art restoration quality alongside remarkable inference efficiency.

\bibliographystyle{IEEEtran}
\bibliography{main}

\end{document}